\documentclass[journal,transmag]{IEEEtran}
\usepackage{times}
\usepackage{epsfig}
\usepackage{graphicx}
\usepackage{amsmath}
\usepackage{amssymb}
\usepackage{bbding}
\usepackage{subfigure}
\usepackage{xcolor}
\usepackage{flushend}
\usepackage{threeparttable}
\usepackage[pagebackref=true,breaklinks=true,letterpaper=true,colorlinks,bookmarks=false]{hyperref}

\usepackage{expl3}
\ExplSyntaxOn
\newcommand\latinabbrev[1]{
	\peek_meaning:NTF . {
		#1\@}%
	{ \peek_catcode:NTF a {
			#1.\@ }%
		{#1.\@}}} 
\ExplSyntaxOff

\def\ie{\latinabbrev{i.e}}

\newcommand{\fig}[1]{Figure~#1}

\newcommand{\sect}[1]{Section~#1}

\begin{document}

\title{Video Instance Segmentation by Instance Flow Assembly}

\author{\IEEEauthorblockN{Xiang Li,
Jinglu Wang,
Xiao Li, 
Yan Lu, \IEEEmembership{Senior Member,~IEEE}
}
\thanks{
This work was done when Xiang Li was an intern at Microsoft Research Asia.
Xiang Li is currently with Department of Electrical and Computer Engineering,
Carnegie Mellon University, Pittsburgh, PA 15213 USA. (email: xl6@andrew.cmu.edu)
}
\thanks{Jinglu Wang, Xiao Li and Yan Lu are with Microsoft Research Asia, Beijing, China.  (email: jinglwa@microsoft; xili11@microsoft.com; yanlu@microsoft.com)
}
}

\markboth{}
{Shell \MakeLowercase{\textit{et al.}}: Bare Demo of IEEEtran.cls for IEEE Transactions on Magnetics Journals}

\IEEEtitleabstractindextext{%
\begin{abstract}
Instance segmentation is a challenging task aiming at classifying and segmenting all object instances of specific classes. While two-stage box-based methods achieve top performances in the image domain, they cannot easily extend their superiority into the video domain. This is because they usually deal with features or images cropped from the detected bounding boxes without alignment, failing to capture pixel-level temporal consistency. 
We embrace the observation that bottom-up methods dealing with box-free features could offer accurate spacial correlations across frames, which can be fully utilized for object and pixel level tracking. 
We first propose our bottom-up framework equipped with a temporal context fusion module to better encode inter-frame correlations. Intra-frame cues for semantic segmentation and object localization are simultaneously extracted and reconstructed by corresponding decoders after a shared backbone. For efficient and robust tracking among instances, we introduce an instance-level correspondence across adjacent frames, which is represented by a center-to-center flow, termed as instance flow, to assemble messy dense temporal correspondences. Experiments demonstrate that the proposed method outperforms the state-of-the-art online methods (taking image-level input) on the challenging Youtube-VIS dataset~\cite{yang2019video}.
\end{abstract}

}

\maketitle

\IEEEdisplaynontitleabstractindextext

\IEEEpeerreviewmaketitle


\section{Introduction}
\label{sec:intro}

Video instance segmentation~\cite{yang2019video}, targeting at classifying, segmenting and tracking object instances of specific classes, is a very challenging task. It finds a lot of applications in video scenarios, such as video editing, autonomous driving and augmented reality.
Recent state-of-the-art methods \cite{bertasius2020classifying,chen2019hybrid,athar2020stem} achieve high segmentation and tracking performance by exploiting video-level cues when taking entire video sequences as inputs. However, such offline methods suffer from high computational cost and intolerable latency. Fast online methods taking image-level inputs are in great demand for practical applications.

\begin{figure}[t]
\centering
\includegraphics[width=\linewidth]{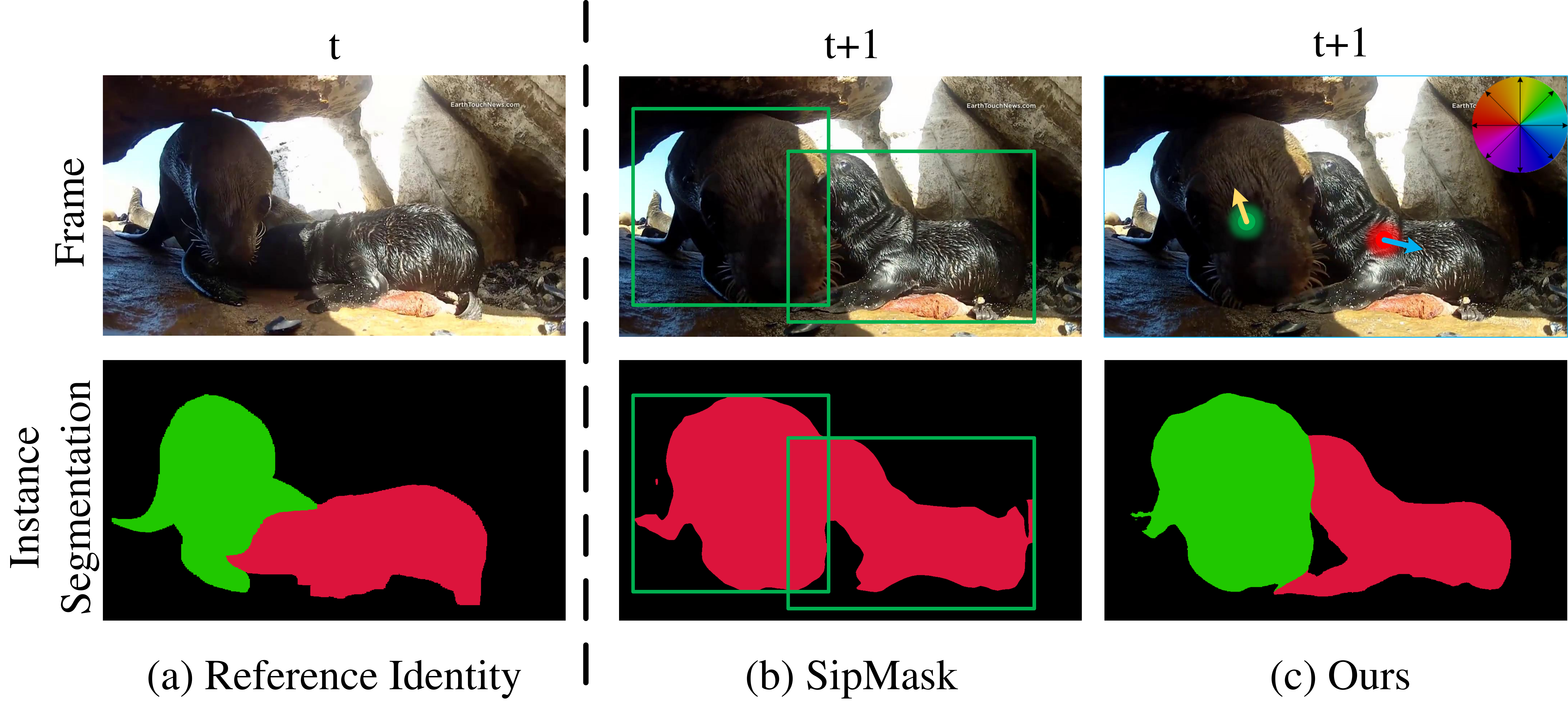}
\caption{Video instance segmentation for similar close instances using the state-of-the-art box-based method (SipMask \cite{Cao_SipMask_ECCV_2020}) and our approach. (a) Ground truth instance identities of the reference frame. (b) SipMask fails to track instance identity from the reference. (c) Our bottom-up approach handles the similar close instances well with a new tracking strategy by preserving spatial correlation based on instance-level flows. Note that the color wheel in the top row indicates the direction of instance-level flows.}
\label{fig:teaser}
\end{figure}

Prevalent image instance segmentation methods, such as Mask-RCNN~\cite{he2017mask}, follow a two-stage paradigm~\cite{liu2018path,chen2019hybrid,huang2019mask,fang2019instaboost}, \ie, first detecting bounding boxes of instances and then segmenting out the foreground inside each box. These image methods can be extended to the video counterpart by adding a tracking component based on the boxes~\cite{Cao_SipMask_ECCV_2020,yang2019video}. Matching or linking instances across frames is implemented by comparing detected bounding boxes or cropped features. 
Specifically, feature similarity, detection confidence and interaction of union (IoU) between bounding boxes are often used. Depending on the detected boxes heavily, these methods have inevitable limitations for cases where multiple similar instances appear, especially when they are close. See the example in \fig{\ref{fig:teaser}} (b) where the seal instances are mis-matched to the ones in the last frame. Erroneous matching occurs because 1) the lack of spatial cues in the cropped features 2) ambiguous feature embedding due to overlapped boxes.

Bottom-up box-free methods also attract increasing research attention in the image domain~\cite{cheng2020panoptic,xie2020polarmask,wang2020solo,wang2020solov2,wang2020axial,yang2019deeperlab,chen2020naive,wang2020max} thanks to faster speed.
Besides, bottom-up methods intrinsically preserves spacial correlation and context and hence it could offer and propagate more cues for instance matching across frames. These advantages make bottom-up methods suitable for video tasks.
Yet, the major challenge for bottom-up image methods exists in assembling low-level semantic and spatial cues. For video instance segmentation, assembling such image cues as well as temporal correlations for instances is the key to improve the performance of bottom-up methods.

In this paper, we present a bottom-up video instance segmentation method by assembling instance-level flow to achieve both high-quality segmentation and robust tracking. We build up the framework based on the image method~\cite{cheng2020panoptic}, and equip it with a temporal feature fusion module to encode inter-frame correlations. To better construct inter-frame correlations, we further introduce a new instance correspondence representation, \ie, instance flow, which is elaborately designed to assembling 2D flow vectors at the instance level. This representation allows to preserve spatial correlation and context of instances and also avoids redundant instance-level feature matching process. 
Taking multiple reference frames into consideration is critical to the robustness of tracking due to occlusion, disocclusion and motion blur. Accordingly, we propose a fast parallelizable assembling method that aggregates instance flows across a set of reference frames. 
We evaluate our method on the challenging video instance segmentation datasets, Youtube-VIS \cite{yang2019video}, against the state-of-the-art track-by-detect methods \cite{Cao_SipMask_ECCV_2020,yang2019video,voigtlaender2019feelvos,wojke2017simple,yang2018efficient}, and state-of-the-art offline methods, \cite{bertasius2020classifying,athar2020stem}. Our method achieves the state-of-the-art result of 34.1 mAP on Youtube-VIS validation set.

In summary, our contributions are three-fold.
\begin{itemize}
    \item A new bottom-up video instance segmentation model equipped with multi-scale temporal context fusion modules, achieving superior performance over most online methods on Youtube-VIS dataset.
    \item An efficient matching representation, instance flow, for encoding inter-frame correspondences between instances in two frames into a 2D vector field.
    \item A parallelizable instance flow assembling method, allowing fast and robust matching by taking advantage of multiple reference frames.
\end{itemize}

\section{Related Works}
\subsection{Image Instance Segmentation.}
Different from semantic segmentation, instance segmentation not only assigns each pixel a semantic class but also an object identity. Existing instance segmentation methods either adopt top-down \cite{he2017mask,Cao_SipMask_ECCV_2020,liu2018path,huang2019mask} or bottom-up \cite{cheng2020panoptic,xie2020polarmask,wang2020solo,wang2020solov2,wang2020axial,yang2019deeperlab,chen2020naive,wang2020max} paradigm. Typically, modern top-down methods rely on a box-based approach that detects object bounding boxes and predicts a mask for each box. Unlike top-down methods, bottom-up methods can represent each instance in a box-free fashion.

In top-down method, instance identity is represented by the bounding-box of detected object. Among those box-based methods, Mask R-CNN \cite{he2017mask} employs a region proposal generation network (RPN) equipped with RoIAlign feature pooling strategy and a feature pyramid networks (FPN) \cite{lin2017feature} to obtain fixed-sized features of each proposal. The pooled features are further used for bounding box prediction and mask segmentation. Followed by Mask R-CNN, several methods are proposed to improve pooling and confidence scoring strategy \cite{li2017fully,liu2018path,huang2019mask}.

In bottom-up methods, there are more flexible ways to represent an instance, such as discriminating instances by their object centers \cite{cheng2020panoptic}, segment each instance mask by calculating object-specific coefficients for corresponding mask prototypes \cite{Cao_SipMask_ECCV_2020}. SOLO \cite{wang2020solo,wang2020solov2} represents each instance by its location and size, thus avoiding discriminating different instances by box. PolorMask \cite{xie2020polarmask} directly models instance contour by using 36 uniformly-spaced rays in polar coordinates, which can be assumed as a generalization of box representation that models each instance contour by 4 rays in polar coordinates. Panoptic-Deeplab \cite{cheng2020panoptic} models semantic segmentation, object center heatmap and center offset (defined as a 2D vector field pointing from each point to its corresponding object center) separately. After that, a computational efficient assembly method is leveraged to correspond each pixel to its correspond instance center and form the final instance segmentation output. SipMask \cite{Cao_SipMask_ECCV_2020} follows the previous one-stage method FCOS \cite{tian2019fcos} and represents each instance by object-specific coefficients and corresponding mask prototypes. To achieve more accurate segmentation result, SipMask divides the input image to four parts and predicts masks in each part separately.

\begin{figure*}[t]
\centering
\includegraphics[width=\textwidth]{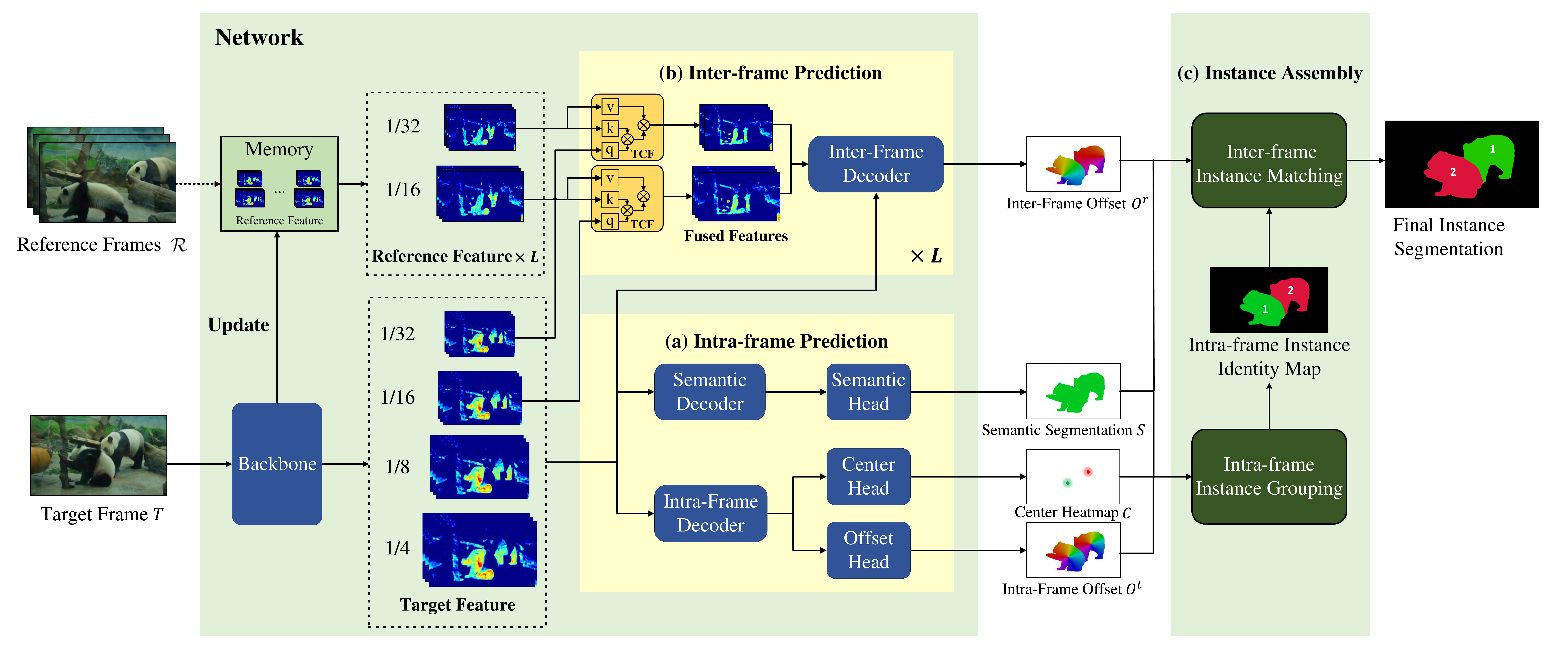}
\caption{\textbf{Method overview}. 
Given a target frame and a set of reference frames, our method outputs instance segmentation result of target frame with consistent identity across frames. Our network contains a backbone, an intra-fame prediction module and an inter-frame prediction module. (a) The intra-frame module predicts semantic segmentation, center heatmap and intra-frame offset, and (b) the inter-frame module equipped with temporal context fusion (TCF) modules predicts inter-frame offset for each reference frame. We perform tracking with (c) instance assembly. The final identities are corrected through inter-frame matching.
}
\label{fig:overview}
\vspace{-0.5cm}
\end{figure*}

\subsection{Video Instance Segmentation.}
Video instance segmentation \cite{yang2019video} is a recently introduced problem which requires classifying, segmenting and tracking object instances across frames. To track instance throughout time, the methods proposed for VIS task work either in an online or offline manner. Typically, Online methods segment instance in each frame then associating them using rules while offline methods directly model the spatial-temporal information of a instance by taking the entire video as inputs. The online methods are more practical for streamlining applications, but the performance of existing methods are far from that of offline methods because of the lack of long-term temporal correlation modeling. 

For online VIS methods \cite{yang2019video, Cao_SipMask_ECCV_2020}, state-of-the-art methods typically leverage a top-down box-based framework and match instance identities regarding to appearance, position and class cues. MaskTrack R-CNN \cite{yang2019video} extends the Mask R-CNN \cite{he2017mask} with a tracking head to extract appearance features of each instance and tracks instances throughout frames by leveraging a combination of position, class and appearance similarity cues. Sip-Mask-VIS \cite{Cao_SipMask_ECCV_2020} follows the tracking strategy used in MaskTrack R-CNN while predicts basis mask and a set of coefficients to improve the segmentation quality. Since it follows the one-stage method FCOS \cite{tian2019fcos} to predict masks, SipMask-VIS is much faster than MaskTrack R-CNN which builds on two-stage Mask R-CNN \cite{he2017mask}.

For offline methods \cite{bertasius2020classifying, athar2020stem, wang2021end}, it typically models the temporal information by directly taking a sequence of frames as inputs. Different from the aforementioned two track-by-detect methods, STEm-Seg \cite{athar2020stem} takes advantage of the sequence-level input and directly models the spatial-temporal relation in the given sequence. More recently, MaskProp \cite{bertasius2020classifying} leverages Mask R-CNN \cite{he2017mask} to conduct image-level instance segmentation then propagate the instance features throughout the sequence to refine masks and propagate instance identities.  More recently, VISTR \cite{wang2021end} utilizes a transformer-based network to model a video clip directly. In particular, a 3D positional encoding is proposed for video prediction tasks. The instance segmentation predictions of VISTR are supervised to ordered thus no post-processing is needed to obtain the continuous instance identities across time.  

\subsection{Video Object Segmentation.}
Video object segmentation (VOS) targets at segmenting foreground instances in a class-agnostic fashion \cite{yang2018efficient,jain2017fusionseg,tokmakov2017learning}. Typically, the ground-truth mask of the first frame is available in this task. 

Some previous methods for semi-supervised VOS rely on fine-tuning at real-time. OSVOS \cite{caelles2017one} leverages an online adaptation mechanism to fine-tune the first-frames ground-truth. MaskTrack \cite{perazzi2017learning} propagates the segmentation masks from one frame to next using optical flow. Although those methods achieves promising results, the inference speed is severely slowed down due to real-time fine-tuning.

Some other recent works aim to avoid online fine-tuning and leads to better run-time. Visual similarity \cite{caelles2017one,chen2018blazingly,yang2020collaborative}, motion cues \cite{cheng2017segflow}, and temporal consistency \cite{perazzi2017learning,yang2018efficient} are extensively used to track and segment the same object throughout the video. Ranking Attention Network (RANet) \cite{wang2019ranet} leverages an encoder-decoder framework to learn pixel-level similarity and adopts a ranking attention module to select feature maps for fine-grained segmentation. Space-Time-Memory networks (STM) \cite{oh2019video} constructs a memory bank for each object in the video and matches each frame to memory bank to gather the temporal correspondence. The newly frames can be added to the memory bank then the bank updates throughout time. Several follow-up works adopt STM as backbone and either apply at other tasks \cite{cheng2021modular, oh2020space}, improve the augmentation policy or training data \cite{cheng2021modular, seong2020kernelized}, modify the memory fetching process \cite{lu2020video, cheng2021modular, seong2020kernelized, li2020fast, hu2021learning}.

\section{Method}

\subsection{Overview}
Given the target frame $T$ and a sequence of reference frames $\mathcal{R} =\{R_i\}, i=1,...,L$, our goal is to densely track object of specific classes with consistent identities. Note that we only take frames before target frames the reference, and thus our method could be performed screamingly without looking ahead. \fig{\ref{fig:overview}} illustrates the overview of our method. We maintain a memory to store frame features computed by a backbone network. In the intra-frame prediction module, inspired by the image-based method \cite{cheng2020panoptic}, the decoders predict a semantic segmentation map $S$, an instance center map $C$ and an intra-frame offset map $O^t$ in the target frame $T$. Each pixel in $O^t=\{(i,j)-c_{ij}^t\}$ stores the offset from its pixel coordinate $(i,j)$ to the assigned instance center coordinate $c_{ij}^t$. To leverage temporal information, we introduce a temporal context fusion module in the inter-frame prediction stage and generate an inter-frame offset map $O^r=\{(i,j)-c_{ij}^r\}$ towards corresponding centers $c_{ij}^r$ in reference frame $R$ for the later instance matching.
Unlike box-based methods \cite{yang2019video} that often using feature similarity for matching, our method assembles bottom-up flows to capture positional and directional information for robust instance matching.
We elaborate detailed network design in \sect{\ref{sec:network}} and instance matching with flows in \sect{\ref{sec:assembly}}.

\subsection{Network Design}
\label{sec:network}
Our network contains three main modules, a shared backbone for frame feature extraction, an intra-frame prediction module to obtain semantic and instance segmentation for current target frame (\fig{\ref{fig:overview}} (a)), and an inter-frame prediction module for fusing temporal features and performing instance matching (\fig{\ref{fig:overview}} (b)).

The backbone architecture of Deeplab V3+ \cite{chen2018encoder} is adopted in our network. To efficiently boost inference speed, we archive the extracted high-level feature map into an external memory. In this way, the required feature maps of reference frame can be directly read from the external memory. 

\subsubsection{Intra-frame Prediction}
Our method adapts the decoder architecture used in \cite{cheng2020panoptic,qiao2020vip} for intra-frame instance segmentation. We leverage decoders paired with separate atrous spatial pyramid pooling (ASPP) \cite{chen2018encoder} for semantic and instance segmentation separately given the condition that those two tasks require different contextual and decoding information. The semantic decoder followed by a semantic head outputs the semantic segmentation $S$. The intra-frame decoder followed by a center head and offset head outputs the center heatmap $C$ and intra-frame offset $O^t$ respectively. In particular, we predict semantic segmentation of objects in various sizes with features in different decoder layers like extensively used FPN \cite{lin2017feature}. It helps to capture multi-scale information thus produces better foreground masks in scenarios, such as partial occlusion and large-scale.

\subsubsection{Inter-frame Prediction}
Our method constructs the inter-frame instance correspondence by regressing the offset $O^r$ towards its corresponding center in reference frames.

\textbf{Temporal context fusion.}
To provide inter-frame information for the decoder, we propose a temporal context fusion (TCF) module to fuse the high-level feature maps between target and reference frames. TCF module takes advantage of the compacted feature maps of both target frame and reference frame as input and works following the standard dot-product attention scheme while being inserted into the network in a pyramid fashion. As shown in \fig{\ref{fig:overview}}, the TCF modules are inserted into the middle of the first two skip connections between encoder and decoder. 

Concerning the network design for inter-frame offset prediction, we empirically add a separate branch in the decoder. 
Sharing the inter-frame decoder with intra-frame modules could fail, because fused features containing position encoding of multiple frames will confuse the network and impair the precision of center prediction.

\subsubsection{Loss Functions}
The total loss $L_{total}$ can be boiled down to five components, cross entropy loss $L_{sem}$ for semantic segmentation, mean squared error loss $L_{cent}$ for instance center prediction, and L1 loss $L_{intra}$, $L_{inter}$ for intra-frame and inter-frame offset regression and shape consistency loss $L_{shape}$, respectively, taking the form:
\begin{equation}
\label{L_total}
\begin{aligned}
\mathcal{L}_{total}=\mathcal{L}_{sem}+&\lambda_{cent}\mathcal{L}_{cent}+\lambda_{inter}\mathcal{L}_{inter}+\\
&\lambda_{intra}\mathcal{L}_{intra}+\lambda_{shape}\mathcal{L}_{shape}
\end{aligned}
\end{equation}

\textbf{Shape consistency loss.}
It is essential to keep the intra-frame and inter-frame offset having consistent shape. Given ground-truths and predictions of intra-frame and inter-frame offset, we calculate the $\mathcal{L}_{shape}$ as Equation~\ref{L_shape}
\begin{equation}
\label{L_shape}
\mathcal{L}_{shape}=\|(\widetilde{O}^r-\widetilde{O}^t)-(O^r-O^t)\|_2^2
\end{equation}
where $\widetilde{O}^t$ and $\widetilde{O}^r$ are the ground-truth of intra-frame and inter-frame offset respectively. And $O^t$ and $O^r$ for predictions of intra-frame and inter-frame offset respectively.

\subsection{Instance Flow Assembly}
\label{sec:assembly}
To robustly track instances across frames, we propose an efficient instance-level correspondence representation, \ie, instance flow, and its assembly strategy. As illustrated in \fig{\ref{fig:overview}} (c), during instance assembly, we first conduct intra-frame instance grouping to obtain the intra-frame identity map, then match and propagate the identity from reference frame to target frame using instance flow. Note that our inter-frame matching process involves multiple reference frames to enhance the tracking robustness.

\begin{figure}[t]
\includegraphics[width=\columnwidth]{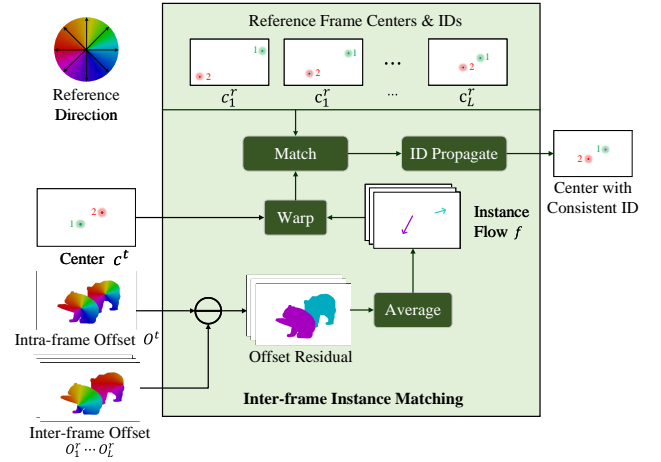}
\caption{\textbf{Instance matching}. The matched instance is selected from the reference frames based on distance between warped center $\bar{c}^t=c^t + f$ and reference center $c^r$ w.r.t. instance flow $f$. The final instance identity is propagated from the matched reference instance to keep the instance identity consistent throughout the sequence.}
\label{IDProp}
\end{figure}

\begin{table*}[!t]
\centering
\caption{Quantitative comparison to the state-of-the-art methods on Youtube-VIS val set.
}
\scalebox{1.5}
{
\begin{tabular}{l|c|l|llll} 
\hline
\hline
Method & Input Type & AP & AP50 & AP75 & AR1 & AR10\\
\hline
\multicolumn{7}{c}{Box-based Methods} \\
\hline
IoUTracker+ \cite{yang2019video} & Image & 23.4 & 36.5 & 25.7 & 28.9 & 31.1\\
SeqTracker \cite{yang2019video} & Image & 27.5 & 45.7 & 28.7 & 29.7 & 32.5\\
DeepSORT \cite{wojke2017simple} & Image & 26.1 & 42.9 & 26.1 & 27.8 & 31.3\\
MaskTrack R-CNN \cite{yang2019video} & Image & 30.3 & 51.1 & 32.6 & 31.0 & 35.5\\
SipMask \cite{Cao_SipMask_ECCV_2020} & Image & 33.7 & 54.1 & 35.8 & 35.4 & 40.1\\
VISTR \cite{wang2021end} & Video & 36.2 & 59.8 & 36.9 & 37.2 & 42.4\\
MaskProp \cite{bertasius2020classifying} & Video & 40.0 & - & 42.9 & - & - \\
\hline
\multicolumn{7}{c}{Box-free Methods} \\
\hline
FEELVOS \cite{voigtlaender2019feelvos}  & Image & 26.9 & 42.0 & 29.7 & 29.9 & 33.4\\
OSMN \cite{yang2018efficient}  & Image & 27.5 & 45.1 & 29.1 & 28.6 & 33.1\\
STEm-Seg \cite{athar2020stem} & Video & 30.6 & 50.7 & 33.5 & 31.6 & 37.1\\
\textbf{Ours} & Image & 34.1 & 54.0 & 37.6 & 36.1 & 41.9\\
\hline
\hline
\end{tabular}
}
\vspace{0.2cm}
\begin{tablenotes}
\item We present the results that are all generated with ResNet-50 as backbone for fairness. The input type denotes the input processed by the system at once.
\end{tablenotes}
\label{main result}
\end{table*}

\textbf{Instance flow.}
Inspired by popular center-based bottom-up instance segmentation methods \cite{wang2020solo,wang2020solov2,cheng2020panoptic,chen2020naive,qiao2020vip}, we learn to segment object instances in videos in a bottom-up fashion by leveraging center representation. Each instance can be represented by a unique geometric center of the instance mask. Thereby, tacking the instances in a video sequence can be converted to tracking the instance centers in the sequence, which is a much easier task. To this end, we define a center-to-center flow, termed instance flow, which can be inferred by the offset residual. For each instance $m$, we denote its center prediction as $c_m$, intra-frame offset as $O^t_m$ and inter-frame offset prediction as $O^r_m$. In practice, we approximate instance flow by sampling and averaging over uniformly-spaced values of offset residual $(O^r_m-O^t_m)$. The instance flow for instance $m$ takes the form
\begin{equation}
\label{instFlow}
f_m = \frac{\iint_{\Omega_m}(O^r_m-O^t_m)\:\mathrm{d}i \mathrm{d}j}{\iint_{\Omega_m}\:\mathrm{d}i \mathrm{d}j }
\end{equation}
where $\Omega_m=\{(i,j) \:|\: \mathtt{ID}_{i,j}=m\}$. Specifically, $\mathtt{ID}$ is a identity map, each pixel $(i,j)$ stores the instance identity it is assigned.

\subsubsection{Intra-frame Instance Grouping}
To obtain the instance identity map for instance flow computing, we assemble the center heatmap $C$, intra-frame offset $O^t$, and semantic segmentation $S$ by a simple grouping operation. we first remove values in background area of heatmap $C$ then obtain valid centers $\{c_m\}, m=\{1,...,M\} $ by performing a pixel-wise non-maximum suppression (NMS) \cite{cheng2020panoptic}. Then, we assign the closest center to each pixel $(i,j)$ by comparing the distances between valid centers $\{c_m\}$ and warped center coordinate $\bar{c}_{ij} = (i,j) + O_{ij}^t$ with respect to offset $O_{ij}^t$. Note that each center represents an unique instance, and thus the instance identity for each pixel $(i,j)$ is given by
\begin{equation}
\label{groupInst}
\mathtt{ID}_{i,j} = \mathop{\arg\min}_{m \in \{1,...,M\} } \|c_m - \bar{c}_{ij})\|_2
\end{equation}
Consequently, pixels are grouped by instance identities. To obtain final intra-frame instance identity map result, we further employ semantic prediction to filter out the background area and keep the instance identity in it as 0. 

\subsubsection{Inter-frame Instance Matching}
Figure~\ref{IDProp} illustrates the matching scheme based on instance flow. Note that the matching process can be easily implemented in a parallel fashion with each target and reference frame pair, and thus the inference time will not increase proportionally when considering multiple reference frames.

\textbf{Instance distance metric.}
To match instance between a reference and target frame pair, we adopt a distance metric to measure the pair-wise instance correspondence. Given a target frame $T$ and a reference frame $R$, the center distance between each instance-pair in $T$ and $R$ can be represented as a matrix $\mathbf{D}_{T,R} \in \mathbb{R}^{M^t\times M^r}$ where $M^t$ and $M^r$ are the number of instances in target frame $T$ and reference frame $R$ respectively. We define each element in $\mathbf{D}_{T,R}$ represents the distance between the warped target center $\bar{c}_m^t = c_m^t + f_m$ with respect to instance flow $f_m$ in $T$ and the reference center $c_n^r$ in frame $R$ by
\begin{equation}
\label{affinity}
d_{m,n} = \|\bar{c}_m^t - c_n^r\|_2
\end{equation}
where $m \in \{1,...,M^t\}$ and $n \in \{1,...,M^r\}$ are the instance identity for instance in frame $T$ and frame $R$ respectively. 

\textbf{Instance matching and propagation.}
Since the video instance segmentation is very challenging due to occlusions or motion blur, instance matching taking a single reference frame is not robust. We explore multiple reference frames to enhance the tracking robustness. Given the instance $m$ in the target frame $T$, we attempt to find the best-matched instance $match(m)$ who has the minimal instance distance to $m$, for all instances appear reference frames $\mathcal{R}$. 

To propagate instance identity, instances in frame $T$ should either inherit identities from appeared instances in past frames $\mathcal{R}$ or be recognized as a new instance then be assigned a new identity. We use a threshold $\epsilon$ to filter unmatched instances based on center pair distance $D_{T,R}$ then assign them new instance identities. 

\subsubsection{Semantic Label}
As defined in \cite{yang2019video}, video instance segmentation task requires only one semantic label to all instances recognized as the same object in the video. In our method, we involve majority voting to determine the category for each instance. Different from \cite{cheng2020panoptic,qiao2020vip}, we accumulate frame-level soft predictions in each frame and get the hard prediction only after traversing all frames in the sequence. 

\subsubsection{Confidence score.}
To evaluate our instance segmentation results, we assign a confidence score to each instance. Following \cite{cheng2020panoptic}, we combine the confidence of  both semantic segmentation and center heatmap to form the final score, which can be computed as 
\begin{equation}
\label{score}
Score_{inst} = Score_{sem} \times Score_{center} 
\end{equation}
where $Score_{sem}$ is obtained from the average of semantic segmentation predictions within the region belonging to each object, and $S_{center}$ is unnormalized objectness score obtained from the class-agnostic center point heatmap.

\begin{figure*}[!t]
\centering
\includegraphics[width=\linewidth]{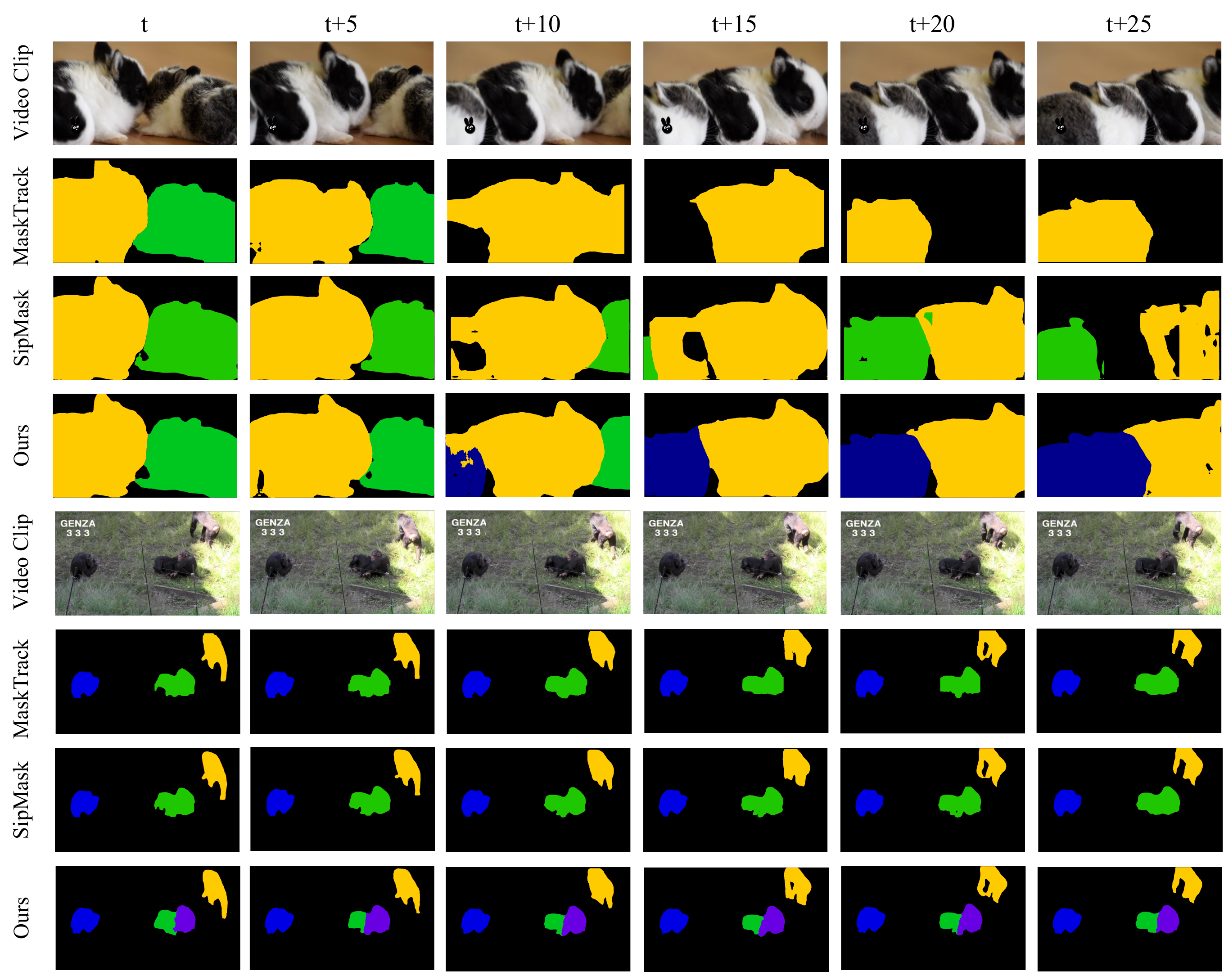}
\caption{Visual comparison to the state-of-the-art video instance segmentation methods, MaskTrack R-CNN \cite{yang2019video} and SipMask \cite{Cao_SipMask_ECCV_2020}. Colors indicate instance identities (best viewed in color).
We show that our method generates more accurate and temporally consistent results, while MaskTrack R-CNN and SipMask attempt to mis-match or even miss instances for the cases where similar instances are close.}
\label{vis2}
\end{figure*}

\section{Experiments}
In this section, we present our results on YouTube-VIS \cite{yang2019video} dataset, which contains 2238 training, 302 validation and 343 test video clips. Each video is annotated with per-pixel segmentation, category, and instance labels. The dataset contains 40 object categories.
Since the evaluation on the test set is currently closed, we perform our evaluations on the validation set. The evaluation metric for this task is defined as the area under the precision-recall curve with different IoUs as thresholds.

\subsection{Implementation Details}
\label{sec:implementation}
\textbf{Training.}
We implement our method on the Pytorch platform. Following Panoptic-Deeplab~\cite{cheng2020panoptic}, we use modified Deeplab-V3+ architecture with Resnet-50 \cite{he2016deep} as the backbone for our network. During training, we adopt a “poly” learning rate policy where the learning rate is multiplied by $(1-\frac{iter}{iter_{max}})^{0.9}$ for each iteration with an initial learning rate of 0.001 to all experiments. An adam \cite{kingma2014adam} optimizer with $\beta_1=0.9$, $\beta_2=0.999$ and $weight decay=0$ is leveraged. In particular, random cropping, scaling, and flipping are used in our training. We crop the input into 640$\times$640 during training. The scales are set to $0.5, 0.6, 0.7, 0.8, 0.9, 1.0\}$. We pre-trained the backbone, semantic-decoder, and intra-frame decoder on COCO with only overlapped 20 categories for 180k iterations. We utilize the pre-trained model and assign inter-frame decoder with the same pre-trained weights of intra-frame decoder then train our model with 8 × NVIDIA V100 for 90k iterations. The batch size is set to 32 during the training. To balance the magnitude of losses, we set $\mathcal{L}_{cent}=100$ and $\mathcal{L}_{inter}=\mathcal{L}_{intra}=\mathcal{L}_{shape}=0.01$. The reference frame is randomly chosen among all frames in the video for inter-frame predictions. To force the network to pay more attention to small objects (as COCO definition \cite{lin2014microsoft}), we give three times weights for small instances when calculating the loss of center and offset predictions. 

\textbf{Inference.}
In the inference step, we set the number of reference frames to four for each target frame as default. The reference frames are selected from the first frame and three adjacent frames before the target frame. We set NMS kernel size to 41, NMS threshold to 0.15 and leverage the multi-scale fusion. The scales are set to $\{0.5, 0.75, 1.0\}$ during inference as prior methods \cite{cheng2020panoptic,chen2020naive,qiao2020vip}. 

\subsection{Main Result}
In this section, we compare our method with other state-of-the-art video instance segmentation methods.

\begin{figure*}[htbp]
\centering
\includegraphics[width=\linewidth]{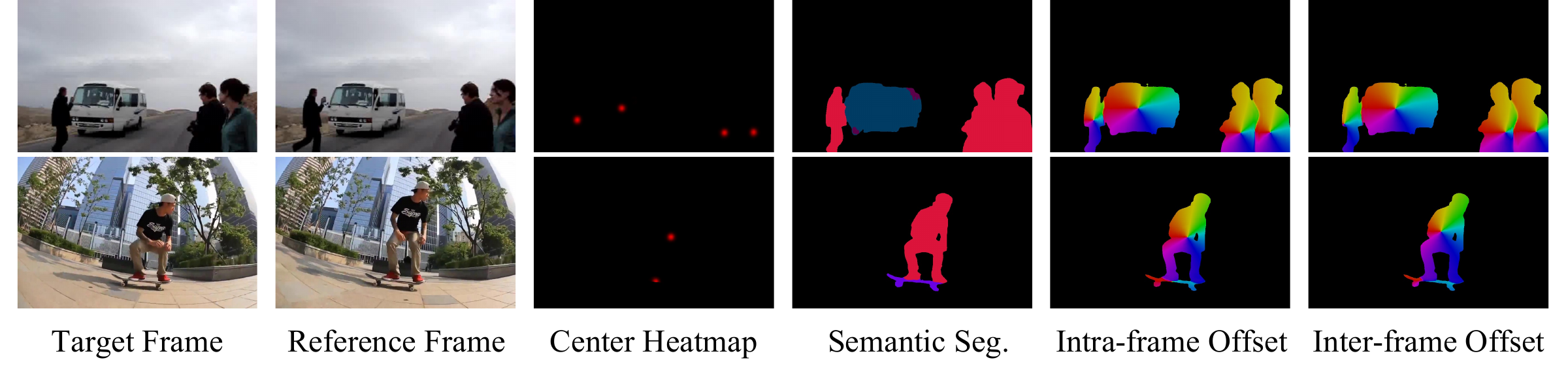}
\caption{Visualization of intermediate predictions of our method. Colors in semantic segmentation indicate category labels. Colors in intra-frame offset and inter-frame offset indicate directions of the offset vectors.}
\vspace{-0.4cm}
\label{fig:intermediate}
\end{figure*}

\begin{figure*}[t]
\centering
\begin{minipage}[t]{0.49\textwidth}
\centering
\includegraphics[width=0.87\linewidth]{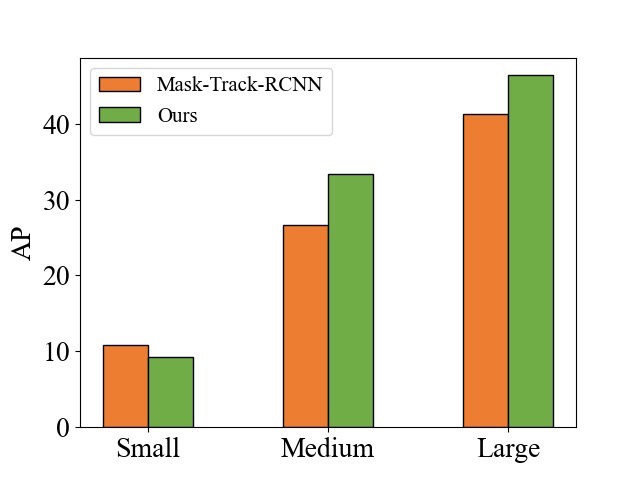}
\end{minipage}
\begin{minipage}[t]{0.49\linewidth}
\centering
\includegraphics[width=0.87\linewidth]{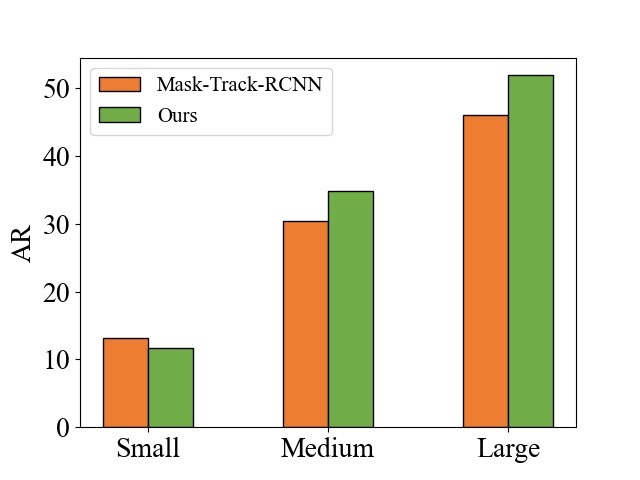}
\end{minipage}
\caption{\textbf{Quantitative comparison with Mask-Track-RCNN \cite{yang2019video} among instances with different sizes.} The AP and AR are average precision and average recall respectively. The small ($area < 32^2$), medium ($32^2 < area < 96^2$), large ($96^2 < area$) instance are defined as the COCO dataset \cite{lin2014microsoft}.}
\label{fig:size_comp}
\end{figure*}

\textbf{Quantitative result.}
We compare our method with popular box-based and box-free video instance segmentation methods in Table~\ref{main result}. Notably, our method achieves an mAP of 34.1 and a real-time speed (37 FPS after parallel optimization on a single NVIDIA V100 16G GPU) given 4 reference frames, which outperforms all other image-level methods, \ie, ones taking image-level input instead of the entire video during inference. Specifically, it surpasses other box-free methods by a large margin of 3.5 percent in terms of mAP. For box-based methods, it outperforms MaskTrack R-CNN by 3.8 and Sip-Mask by 0.4 in terms of mAP. The gap between our method and MaskProp~\cite{bertasius2020classifying} mainly comes from: (1) MaskProp is an offline method that combines multiple networks to propagate features and masks to neighbor frames for refinement. (2) MaskProp involves several strong feature extraction modules such as deformable convolution \cite{dai2017deformable} to align the features. Since we aim to explore efficient online video instance segmentation method, sequence-level correspondence and complicated network structures are beyond the scope of this work.

Figure~\ref{fig:size_comp} illustrates the average precision and average recall of our model against MaskTrack-RCNN \cite{yang2019video}. The segmentation quality decreases with the instance size varying from large to small as the COCO definition \cite{lin2014microsoft}. We found that our method has the better performance for medium and large instances in terms of both average precision and average recall while, for small instances, our method shows inferior performance. Two reasons may result in the degradation - (a) the center of small instances are more difficult to predict, (b) since the instance flow is averaged from the offset residual on each object region, the small instance has fewer pixels to average thus perhaps leading to inaccurate instance flow.

\textbf{Qualitative result.}
We present our qualitative result in Figure~\ref{vis2} and compare it against the state-of-the-art top-down methods MaskTrack R-CNN \cite{yang2019video} and SipMask \cite{Cao_SipMask_ECCV_2020}. The result shows that MaskTrack R-CNN and SipMask fail to track or even segment instances for the cases where similar instances are close. In contrast, our method produces high-quality instance segmentation and tracking results. This implies that our method generates more accurate and temporally consistent results than previous box-based methods. To better illustrate our instance assembly process, we also visualize some intermediate results predicted by the network including center heatmap, semantic segmentation, intra-frame offset and inter-frame offset shown in Figure~\ref{fig:intermediate}. To best visualize, we mask both center and offset maps with predicted foreground mask in the semantic segmentation.

\subsection{Ablation Study}
\label{sec:ablation}

\begin{table*}[ht]
\setlength{\fboxrule}{0pt}
\fbox{
\begin{minipage}[t]{\textwidth}
\begin{minipage}[h]{0.47\textwidth}
\makeatletter\def\@captype{table}
\centering
\caption{Ablation experiments for inter-frame feature fusion method.}
\begin{tabular}{ccc} 
\hline
\hline
Fusion Method & mAP & AP75\\
\hline
Concat + CONVS & 30.5 & 32.0\\
Concat + Cascade ASPP & 31.0 & 32.6\\
TCF module & 34.1 & 37.6\\
\hline
\hline
\end{tabular}
\begin{tablenotes}
\item \textbf{Cascade ASPP} \cite{qiao2020vip}: Cascaded Atrous Spatial Pyramid Pooling. 
\item \textbf{CONVS}: 2D-convolution with the same depths of Cascade ASPP.
\end{tablenotes}
\label{ablation:TCF}
\end{minipage}
\hspace{0.04\textwidth}
\begin{minipage}[h]{0.47\textwidth}
\makeatletter\def\@captype{table}
\centering
\caption{Ablation experiments for reference frame number. 
}
\begin{tabular}{ccc} 
\hline
\hline
Reference Frame & mAP & AP75\\
\hline
1 adjacent frame & 26.2 & 25.4\\
First + adjacent 1 frame & 28.0 & 28.9\\
First + adjacent 2 frames & 33.4 & 37.4\\
First + adjacent 3 frames & 34.1 & 37.6\\
First + adjacent 4 frames & 32.2 & 36.1\\
\hline
\hline
\end{tabular}
\begin{tablenotes}
\item Note that we use adjacent frames before the target frame.
\end{tablenotes}
\label{ablation:reference frame number}
\end{minipage}
\end{minipage}}
\vspace{-0.3cm}
\end{table*}

\begin{table*}[ht]
\setlength{\fboxrule}{0pt}
\scalebox{1}{
\begin{minipage}[t]{\textwidth}
\centering
\begin{minipage}[t]{0.47\textwidth}
\label{flow estimation}
\centering
\makeatletter\def\@captype{table}
\caption{Ablation experiments for flow estimation methods.}
\begin{tabular}{ccc} 
\hline
\hline
Offset Category & mAP & AP75\\
\hline
Affinity field \cite{cao2019openpose} & 23.3 & 24.1\\
Inter-frame offset-IoU\cite{qiao2020vip} & 26.0 & 27.0\\
Inter-frame offset-AVG & 30.4 & 31.4\\
Offset residual (\textbf{Ours}) & 32.6 & 36.0\\
Offset residual + shape loss (\textbf{Ours}) & 34.1 & 37.6\\
\hline
\hline
\end{tabular}
\begin{tablenotes}
\item Three instance correspondence representations are compared, including affinity field, inter-frame offset and offset residual. To fairly compared with our method, two strategies are leveraged to compute the instance flow using inter-frame offset.
\end{tablenotes}
\label{ablation:flow estimation}
\end{minipage}
\hspace{0.04\textwidth}
\begin{minipage}[t]{0.47\textwidth}
\centering
\makeatletter\def\@captype{table}
\caption{Ablation experiments of different training settings of background supervisions.}
\begin{tabular}{cccccc} 
\hline
\hline
Center & Intra-offset & Inter-offset & mAP & AP75\\
\hline
\Checkmark & & & 32.2 & 34.3\\
& \Checkmark & \Checkmark & 29.2 & 30.6\\
\Checkmark &\Checkmark & \Checkmark & 28.1 & 28.1\\
& & & 34.1 & 37.6\\
\hline
\hline
\end{tabular}
\begin{tablenotes}
\item $\checkmark$ here denotes training with loss in background area.
\end{tablenotes}
\label{ablation:background}
\end{minipage}
\end{minipage}}
\vspace{-0.3cm}
\end{table*}

\begin{table*}[ht]
\setlength{\fboxrule}{0pt}
\scalebox{1}{
\begin{minipage}[t]{\textwidth}
\centering
\begin{minipage}[t]{0.47\textwidth}
\centering
\makeatletter\def\@captype{table}
\caption{Ablation experiments for instance scoring.}
\begin{tabular}{ccc} 
\hline
\hline
Score & mAP & AP75\\
\hline
Semantic Score & 33.8 & 37.5\\
Center Score & 33.5 & 37.3\\
Instance Score & 34.1 & 37.6\\
\hline
\hline
\end{tabular}
\begin{tablenotes}
\item Semantic score is the averaged value of the class probability of each object. Center score is the peak value of the predicted gaussian center heatmap of each object. Instance score is product of semantic score and center score.
\end{tablenotes}
\label{ablation:score}
\end{minipage}
\hspace{0.04\textwidth}
\begin{minipage}[t]{0.47\textwidth}
\makeatletter\def\@captype{table}
\caption{Ablation experiments of threshold of intra-frame instance grouping.}
\centering
\begin{tabular}{cccc} 
\hline
\hline
Window size & Threshold & mAP & AP75\\
\hline
21 & 0.05 & 32.0 & 36.1\\ 
21 & 0.15 & 33.9 & 37.4\\ 
21 & 0.25 & 33.6 & 36.7\\ 
41 & 0.05 & 32.1 & 36.1\\ 
41 & 0.15 & 34.1 & 37.6\\ 
41 & 0.25 & 33.9 & 37.1\\  
\hline
\hline
\end{tabular}
\begin{tablenotes}
\item The window size and threshold are for NMS of the center heatmap in intra-frame instance grouping.
\end{tablenotes}
\label{ablation:nms}
\end{minipage}
\end{minipage}}
\end{table*}

We conduct extensive ablation experiments on the same backbone (ResNet-50) with several design choices, in terms of network module, reference frames and instance flow.

\begin{figure}[t]
\centering
\includegraphics[width=\linewidth]{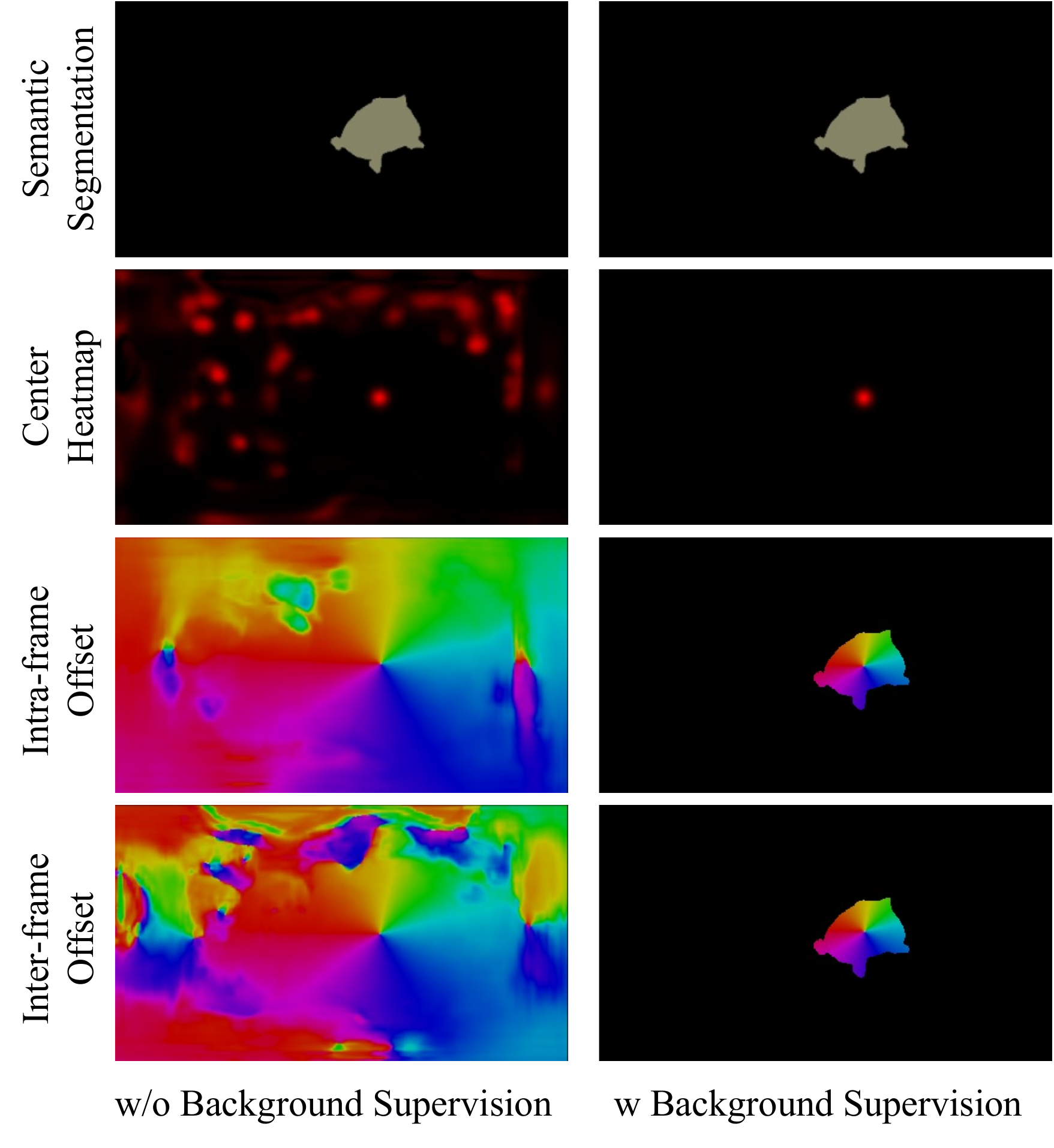}
\caption{\textbf{Visualization of predictions with or w/o loss in background area.} A filter is applied on offset predictions to filter out small values before visualizing. For semantic segmentation and offset predictions, the colors represents classes and directions respectively.}
\label{fig:bgc_comp}
\vspace{-0.3cm}
\end{figure}

\textbf{Inter-frame feature fusion.}
To investigate the effectiveness of our TCF module, we train two baseline models with the same setting as our best result. As shown in Table~\ref{ablation:TCF}, the TCF module brings 3.6 mAP and 3.1 mAP gain against pure convolutions and cascade ASPP \cite{qiao2020vip} respectively. This demonstrates that our proposed TCF module helps the decoder learn better embeddings of inter-frame correspondence. 

\textbf{Reference frame number.}
In Table~\ref{ablation:reference frame number}, we experiment with different reference frame numbers when conducting the inter-frame instance matching. We found that leveraging a single reference frame only leads to 26.2 mAP whilst multiple reference frames matching can produce our best result of 34.1 mAP with a reference frame number of four. Here, the first frame is involved to mitigate the error propagation problem. We would like to highlight that our multi-frame matching strategy is designed to be easily parallelable, thus no additional latency will be introduced by utilizing multiple reference frames in comparison to using only a single reference frame after parallel optimization.

\textbf{Flow assembly methods.}
Instance flow is the core of our assembly method in the tracking procedure. Here, we elaborate on the motivation that we compute it from the offset residual. We perform the comparison on different flow estimation methods taking pure inter-frame offset and affinity field~\cite{cao2019openpose}.

The affinity field is extensively used in human pose estimation while we modify it here to define the affinity field as a unit vector field on instance movement trajectories. During inference, the model trained with affinity field shares a similar instance assembling scheme as inter-frame instance matching module. The only difference is that the instance flow $f$ is calculated by the line integral along the instance movement trajectory as defined in \cite{cao2019openpose}. Besides, since inter-frame offset cannot be used to propagate instance identity directly, we compare our model with two baseline settings. We first compare the IoU-based single-frame propagation method proposed in \cite{qiao2020vip}. Then, to fairly compared with our multi-frame matching pipeline, we calculate the instance flow by directly averaging the inter-frame offset separately for each instance given the condition that the regressed object centers are defined as the instance barycentric centers. The results are shown in Table~\ref{ablation:flow estimation} implies that offset residual is the best approach to estimate the instance flow. 

To analyze, different from affinity field defined on limbs in pose estimation, movement trajectories of instances have no semantic meaning thus hard for the network to regress. In addition, the affinity field will be ambiguous facing intersect movement trajectories. Concerning the result of inter-frame offset, the low mAP of 26.0 for single frame IoU matching suggests that single reference frame is not robust in the video instance segmentation task. Similarly, by averaging the inter-frame offset directly, the averaged value equals our defined instance flow only in the condition that both center and foreground predictions are perfect. However, this is not true in challenging scenarios, such as motion blur and occlusion, thus resulting in a lower performance of 30.4 mAP than offset residual. Specifically, offset residual addresses this problem by calculating a relative vector between two offsets. This benefits the instance flow estimation by avoiding the negative influence of incomplete foreground mask predictions. It is worth to mentioning that intentionally constraining the shape consistency between intra-frame and inter-frame offset also benefits our method. Without shape loss, the performance will drop by around 1.5 mAP. 

\textbf{Supervision in background area.}
The background area of offset and center prediction is filtered out during instance assembly. We mask out the loss in the background area during training and expect it to bring better performance. To verify this, we ablate the supervisions with different settings in the background area as presented in Table~\ref{ablation:background}. For the cases with loss in the background area, the ground-truth value in the background is set to zero. As shown in Table~\ref{ablation:background}, training network with background loss for center prediction results in a 1.9 mAP drop compared to our default training scheme. On the other hand, with background offset losses, the performance will plummet to 29.2 mAP. As shown in Figure~\ref{fig:bgc_comp}, the center heatmap and offset predictions have random patterns in the background area. However, the predictions in the  foreground area are clear. The offset prediction has a broader area than the instance pointing to the instance center, which can predict a more accurate offset residual without possible blurred edges. 

The reason that background supervisions result in worse performance may attribute to two reasons. First, the network can save parameters to suppress the values in the background without supervision, which may bring better foreground and edge results. Besides, constraining values in the background area to zero values will bring false-negative predictions thus missing instances. In our experiment, this happens, especially for small instances. Since our method tracks instances depending on previous instance segmentation results, seamless low-quality instance predictions can break our tracking pipeline.

\textbf{Instance scoring.}
VIS task requires to give instance confidence before evaluation. Previous methods always leverage the detection confidence \cite{yang2019video} as the confidence score for instance prediction. However, since VIS assumes an instance as true positive only if it has correct class prediction as well as a mask prediction with over 0.5 IoU of ground-truth, a sole high detection score can not lead to the true positive rate. We combine the semantic confidence and center confidence \cite{cheng2020panoptic} here to consider both class and position aspects. As shown in Table~\ref{ablation:score}, purely utilizing semantic confidence or center confidence will result in sub-optimal performance. While using instance score brings the best result which eclipses that of using the semantic score and center score by 0.3 and 0.6 mAP respectively.

\textbf{NMS parameters of intra-frame instance grouping.}
In this section, we show the influence of different parameter choices when conducting the intra-frame instance grouping. Since we represent each instance by its center, it is vital to get an accurate center coordinate from the gaussian center heatmap. We conduct NMS to filter out mom-maximum values and only keep one peak value in the gaussian heatmap within a given window (equivalent to a max-pooling) as a valid center of instance if it is larger than a threshold. We ablate the window size and threshold value as shown in Table~\ref{ablation:nms}. The window size shows a minor influence on the final instance prediction performance while the center threshold has a relatively stronger impact on results. Since the threshold is applied to select high confident centers, a smaller threshold may result in a slack center selection and create redundant centers in the inter-frame instance matching. In contrast, if a restricted threshold is adopted, some instance centers may be neglected which will result in the missing of instances.

\section{Conclusion}
In this paper, we have presented a bottom-up one-stage framework for video instance segmentation. We impose a dot-product attention-based temporal context fusion module to learn the embeddings of implicit inter-frame instance correspondence. To leverage the learned embeddings, we further introduce the instance flow to encode inter-frame instance correspondence into a 2D vector flow. To facilitate multiple frame instance matching, an efficient multiple frame matching strategy is further proposed. The comprehensive experiments on the Youtube-VIS dataset verify the effectiveness of the proposed approach and show that proposed method outperforms current image-level state-of-the-art works on video instance segmentation task.

\ifCLASSOPTIONcaptionsoff
  \newpage
\fi

{\small
\bibliographystyle{ieee_fullname}
\bibliography{egbib}

\begin{thebibliography}{10}\itemsep=-1pt

\bibitem{athar2020stem}
Ali Athar, Sabarinath Mahadevan, Aljssa Osep, Laura Leal-Taix{\'e}, and Bastian
  Leibe.
\newblock Stem-seg: Spatio-temporal embeddings for instance segmentation in
  videos.
\newblock In {\em European Conference on Computer Vision}, pages 158--177.
  Springer, 2020.

\bibitem{bertasius2020classifying}
Gedas Bertasius and Lorenzo Torresani.
\newblock Classifying, segmenting, and tracking object instances in video with
  mask propagation.
\newblock In {\em Proceedings of the IEEE/CVF Conference on Computer Vision and
  Pattern Recognition}, pages 9739--9748, 2020.

\bibitem{caelles2017one}
Sergi Caelles, Kevis-Kokitsi Maninis, Jordi Pont-Tuset, Laura Leal-Taix{\'e},
  Daniel Cremers, and Luc Van~Gool.
\newblock One-shot video object segmentation.
\newblock In {\em Proceedings of the IEEE conference on computer vision and
  pattern recognition}, pages 221--230, 2017.

\bibitem{Cao_SipMask_ECCV_2020}
Jiale Cao, Rao~Muhammad Anwer, Hisham Cholakkal, Fahad~Shahbaz Khan, Yanwei
  Pang, and Ling Shao.
\newblock Sipmask: Spatial information preservation for fast instance
  segmentation, 2020.

\bibitem{cao2019openpose}
Zhe Cao, Gines Hidalgo, Tomas Simon, Shih-En Wei, and Yaser Sheikh.
\newblock Openpose: realtime multi-person 2d pose estimation using part
  affinity fields.
\newblock {\em IEEE transactions on pattern analysis and machine intelligence},
  43(1):172--186, 2019.

\bibitem{chen2019hybrid}
Kai Chen, Jiangmiao Pang, Jiaqi Wang, Yu Xiong, Xiaoxiao Li, Shuyang Sun,
  Wansen Feng, Ziwei Liu, Jianping Shi, Wanli Ouyang, et~al.
\newblock Hybrid task cascade for instance segmentation.
\newblock In {\em Proceedings of the IEEE/CVF Conference on Computer Vision and
  Pattern Recognition}, pages 4974--4983, 2019.

\bibitem{chen2020naive}
Liang-Chieh Chen, Raphael~Gontijo Lopes, Bowen Cheng, Maxwell~D Collins, Ekin~D
  Cubuk, Barret Zoph, Hartwig Adam, and Jonathon Shlens.
\newblock Naive-student: Leveraging semi-supervised learning in video sequences
  for urban scene segmentation.
\newblock In {\em European Conference on Computer Vision}, pages 695--714.
  Springer, 2020.

\bibitem{chen2018encoder}
Liang-Chieh Chen, Yukun Zhu, George Papandreou, Florian Schroff, and Hartwig
  Adam.
\newblock Encoder-decoder with atrous separable convolution for semantic image
  segmentation.
\newblock In {\em Proceedings of the European conference on computer vision
  (ECCV)}, pages 801--818, 2018.

\bibitem{chen2018blazingly}
Yuhua Chen, Jordi Pont-Tuset, Alberto Montes, and Luc Van~Gool.
\newblock Blazingly fast video object segmentation with pixel-wise metric
  learning.
\newblock In {\em Proceedings of the IEEE conference on computer vision and
  pattern recognition}, pages 1189--1198, 2018.

\bibitem{cheng2020panoptic}
Bowen Cheng, Maxwell~D Collins, Yukun Zhu, Ting Liu, Thomas~S Huang, Hartwig
  Adam, and Liang-Chieh Chen.
\newblock Panoptic-deeplab: A simple, strong, and fast baseline for bottom-up
  panoptic segmentation.
\newblock In {\em Proceedings of the IEEE/CVF Conference on Computer Vision and
  Pattern Recognition}, pages 12475--12485, 2020.

\bibitem{cheng2021modular}
Ho~Kei Cheng, Yu-Wing Tai, and Chi-Keung Tang.
\newblock Modular interactive video object segmentation: Interaction-to-mask,
  propagation and difference-aware fusion.
\newblock In {\em Proceedings of the IEEE/CVF Conference on Computer Vision and
  Pattern Recognition}, pages 5559--5568, 2021.

\bibitem{cheng2017segflow}
Jingchun Cheng, Yi-Hsuan Tsai, Shengjin Wang, and Ming-Hsuan Yang.
\newblock Segflow: Joint learning for video object segmentation and optical
  flow.
\newblock In {\em Proceedings of the IEEE international conference on computer
  vision}, pages 686--695, 2017.

\bibitem{dai2017deformable}
Jifeng Dai, Haozhi Qi, Yuwen Xiong, Yi Li, Guodong Zhang, Han Hu, and Yichen
  Wei.
\newblock Deformable convolutional networks.
\newblock In {\em Proceedings of the IEEE international conference on computer
  vision}, pages 764--773, 2017.

\bibitem{fang2019instaboost}
Hao-Shu Fang, Jianhua Sun, Runzhong Wang, Minghao Gou, Yong-Lu Li, and Cewu Lu.
\newblock Instaboost: Boosting instance segmentation via probability map guided
  copy-pasting.
\newblock In {\em Proceedings of the IEEE/CVF International Conference on
  Computer Vision}, pages 682--691, 2019.

\bibitem{he2017mask}
Kaiming He, Georgia Gkioxari, Piotr Doll{\'a}r, and Ross Girshick.
\newblock Mask r-cnn.
\newblock In {\em Proceedings of the IEEE international conference on computer
  vision}, pages 2961--2969, 2017.

\bibitem{he2016deep}
Kaiming He, Xiangyu Zhang, Shaoqing Ren, and Jian Sun.
\newblock Deep residual learning for image recognition.
\newblock In {\em Proceedings of the IEEE conference on computer vision and
  pattern recognition}, pages 770--778, 2016.

\bibitem{hu2021learning}
Li Hu, Peng Zhang, Bang Zhang, Pan Pan, Yinghui Xu, and Rong Jin.
\newblock Learning position and target consistency for memory-based video
  object segmentation.
\newblock In {\em Proceedings of the IEEE/CVF Conference on Computer Vision and
  Pattern Recognition}, pages 4144--4154, 2021.

\bibitem{huang2019mask}
Zhaojin Huang, Lichao Huang, Yongchao Gong, Chang Huang, and Xinggang Wang.
\newblock Mask scoring r-cnn.
\newblock In {\em Proceedings of the IEEE/CVF Conference on Computer Vision and
  Pattern Recognition}, pages 6409--6418, 2019.

\bibitem{jain2017fusionseg}
Suyog~Dutt Jain, Bo Xiong, and Kristen Grauman.
\newblock Fusionseg: Learning to combine motion and appearance for fully
  automatic segmentation of generic objects in videos.
\newblock In {\em 2017 IEEE conference on computer vision and pattern
  recognition (CVPR)}, pages 2117--2126. IEEE, 2017.

\bibitem{kingma2014adam}
Diederik~P Kingma and Jimmy Ba.
\newblock Adam: A method for stochastic optimization.
\newblock {\em arXiv preprint arXiv:1412.6980}, 2014.

\bibitem{li2017fully}
Yi Li, Haozhi Qi, Jifeng Dai, Xiangyang Ji, and Yichen Wei.
\newblock Fully convolutional instance-aware semantic segmentation.
\newblock In {\em Proceedings of the IEEE conference on computer vision and
  pattern recognition}, pages 2359--2367, 2017.

\bibitem{li2020fast}
Yu Li, Zhuoran Shen, and Ying Shan.
\newblock Fast video object segmentation using the global context module.
\newblock In {\em European Conference on Computer Vision}, pages 735--750.
  Springer, 2020.

\bibitem{lin2017feature}
Tsung-Yi Lin, Piotr Doll{\'a}r, Ross Girshick, Kaiming He, Bharath Hariharan,
  and Serge Belongie.
\newblock Feature pyramid networks for object detection.
\newblock In {\em Proceedings of the IEEE conference on computer vision and
  pattern recognition}, pages 2117--2125, 2017.

\bibitem{lin2014microsoft}
Tsung-Yi Lin, Michael Maire, Serge Belongie, James Hays, Pietro Perona, Deva
  Ramanan, Piotr Doll{\'a}r, and C~Lawrence Zitnick.
\newblock Microsoft coco: Common objects in context.
\newblock In {\em European conference on computer vision}, pages 740--755.
  Springer, 2014.

\bibitem{liu2018path}
Shu Liu, Lu Qi, Haifang Qin, Jianping Shi, and Jiaya Jia.
\newblock Path aggregation network for instance segmentation.
\newblock In {\em Proceedings of the IEEE conference on computer vision and
  pattern recognition}, pages 8759--8768, 2018.

\bibitem{lu2020video}
Xiankai Lu, Wenguan Wang, Martin Danelljan, Tianfei Zhou, Jianbing Shen, and
  Luc Van~Gool.
\newblock Video object segmentation with episodic graph memory networks.
\newblock In {\em Computer Vision--ECCV 2020: 16th European Conference,
  Glasgow, UK, August 23--28, 2020, Proceedings, Part III 16}, pages 661--679.
  Springer, 2020.

\bibitem{oh2019video}
Seoung~Wug Oh, Joon-Young Lee, Ning Xu, and Seon~Joo Kim.
\newblock Video object segmentation using space-time memory networks.
\newblock In {\em Proceedings of the IEEE/CVF International Conference on
  Computer Vision}, pages 9226--9235, 2019.

\bibitem{oh2020space}
Seoung~Wug Oh, Joon-Young Lee, Ning Xu, and Seon~Joo Kim.
\newblock Space-time memory networks for video object segmentation with user
  guidance.
\newblock {\em IEEE Transactions on Pattern Analysis \& Machine Intelligence},
  (01):1--1, 2020.

\bibitem{perazzi2017learning}
Federico Perazzi, Anna Khoreva, Rodrigo Benenson, Bernt Schiele, and Alexander
  Sorkine-Hornung.
\newblock Learning video object segmentation from static images.
\newblock In {\em Proceedings of the IEEE conference on computer vision and
  pattern recognition}, pages 2663--2672, 2017.

\bibitem{qiao2020vip}
Siyuan Qiao, Yukun Zhu, Hartwig Adam, Alan Yuille, and Liang-Chieh Chen.
\newblock Vip-deeplab: Learning visual perception with depth-aware video
  panoptic segmentation.
\newblock {\em arXiv preprint arXiv:2012.05258}, 2020.

\bibitem{seong2020kernelized}
Hongje Seong, Junhyuk Hyun, and Euntai Kim.
\newblock Kernelized memory network for video object segmentation.
\newblock In {\em European Conference on Computer Vision}, pages 629--645.
  Springer, 2020.

\bibitem{tian2019fcos}
Zhi Tian, Chunhua Shen, Hao Chen, and Tong He.
\newblock Fcos: Fully convolutional one-stage object detection.
\newblock In {\em Proceedings of the IEEE/CVF international conference on
  computer vision}, pages 9627--9636, 2019.

\bibitem{tokmakov2017learning}
Pavel Tokmakov, Karteek Alahari, and Cordelia Schmid.
\newblock Learning video object segmentation with visual memory.
\newblock In {\em Proceedings of the IEEE International Conference on Computer
  Vision}, pages 4481--4490, 2017.

\bibitem{voigtlaender2019feelvos}
Paul Voigtlaender, Yuning Chai, Florian Schroff, Hartwig Adam, Bastian Leibe,
  and Liang-Chieh Chen.
\newblock Feelvos: Fast end-to-end embedding learning for video object
  segmentation.
\newblock In {\em Proceedings of the IEEE/CVF Conference on Computer Vision and
  Pattern Recognition}, pages 9481--9490, 2019.

\bibitem{wang2020max}
Huiyu Wang, Yukun Zhu, Hartwig Adam, Alan Yuille, and Liang-Chieh Chen.
\newblock Max-deeplab: End-to-end panoptic segmentation with mask transformers.
\newblock {\em arXiv preprint arXiv:2012.00759}, 2020.

\bibitem{wang2020axial}
Huiyu Wang, Yukun Zhu, Bradley Green, Hartwig Adam, Alan Yuille, and
  Liang-Chieh Chen.
\newblock Axial-deeplab: Stand-alone axial-attention for panoptic segmentation.
\newblock In {\em European Conference on Computer Vision}, pages 108--126.
  Springer, 2020.

\bibitem{wang2020solo}
Xinlong Wang, Tao Kong, Chunhua Shen, Yuning Jiang, and Lei Li.
\newblock Solo: Segmenting objects by locations.
\newblock In {\em European Conference on Computer Vision}, pages 649--665.
  Springer, 2020.

\bibitem{wang2020solov2}
Xinlong Wang, Rufeng Zhang, Tao Kong, Lei Li, and Chunhua Shen.
\newblock Solov2: Dynamic, faster and stronger.
\newblock {\em arXiv preprint arXiv:2003.10152}, 2020.

\bibitem{wang2021end}
Yuqing Wang, Zhaoliang Xu, Xinlong Wang, Chunhua Shen, Baoshan Cheng, Hao Shen,
  and Huaxia Xia.
\newblock End-to-end video instance segmentation with transformers.
\newblock In {\em Proceedings of the IEEE/CVF Conference on Computer Vision and
  Pattern Recognition}, pages 8741--8750, 2021.

\bibitem{wang2019ranet}
Ziqin Wang, Jun Xu, Li Liu, Fan Zhu, and Ling Shao.
\newblock Ranet: Ranking attention network for fast video object segmentation.
\newblock In {\em Proceedings of the IEEE/CVF International Conference on
  Computer Vision}, pages 3978--3987, 2019.

\bibitem{wojke2017simple}
Nicolai Wojke, Alex Bewley, and Dietrich Paulus.
\newblock Simple online and realtime tracking with a deep association metric.
\newblock In {\em 2017 IEEE international conference on image processing
  (ICIP)}, pages 3645--3649. IEEE, 2017.

\bibitem{xie2020polarmask}
Enze Xie, Peize Sun, Xiaoge Song, Wenhai Wang, Xuebo Liu, Ding Liang, Chunhua
  Shen, and Ping Luo.
\newblock Polarmask: Single shot instance segmentation with polar
  representation.
\newblock In {\em Proceedings of the IEEE/CVF conference on computer vision and
  pattern recognition}, pages 12193--12202, 2020.

\bibitem{yang2019video}
Linjie Yang, Yuchen Fan, and Ning Xu.
\newblock Video instance segmentation.
\newblock In {\em Proceedings of the IEEE/CVF International Conference on
  Computer Vision}, pages 5188--5197, 2019.

\bibitem{yang2018efficient}
Linjie Yang, Yanran Wang, Xuehan Xiong, Jianchao Yang, and Aggelos~K
  Katsaggelos.
\newblock Efficient video object segmentation via network modulation.
\newblock In {\em Proceedings of the IEEE Conference on Computer Vision and
  Pattern Recognition}, pages 6499--6507, 2018.

\bibitem{yang2019deeperlab}
Tien-Ju Yang, Maxwell~D Collins, Yukun Zhu, Jyh-Jing Hwang, Ting Liu, Xiao
  Zhang, Vivienne Sze, George Papandreou, and Liang-Chieh Chen.
\newblock Deeperlab: Single-shot image parser.
\newblock {\em arXiv preprint arXiv:1902.05093}, 2019.

\bibitem{yang2020collaborative}
Zongxin Yang, Yunchao Wei, and Yi Yang.
\newblock Collaborative video object segmentation by foreground-background
  integration.
\newblock In {\em European Conference on Computer Vision}, pages 332--348.
  Springer, 2020.

\end{thebibliography}
}




\end{document}